\documentclass{article}

\PassOptionsToPackage{numbers,compress}{natbib}

\usepackage[preprint]{neurips_2026}

\usepackage[utf8]{inputenc}
\usepackage[T1]{fontenc}
\usepackage{hyperref}
\usepackage{url}
\usepackage{booktabs}
\usepackage{amsfonts}
\usepackage{nicefrac}
\usepackage{microtype}
\usepackage{xcolor}
\usepackage{multirow}
\usepackage{amsmath}
\usepackage{amssymb}
\usepackage{graphicx}
\usepackage{colortbl}
\usepackage{wrapfig}
\usepackage{algorithm}
\usepackage{algpseudocodex}
\usepackage{enumitem}
\usepackage{longtable}

\definecolor{myframepink}{RGB}{72,138,176}
\hypersetup{colorlinks=true,allcolors=myframepink}

\title{LLMs Should Not Yet Be Credited with Decision Explanation}

\author{
Wenshuo Wang \\
School of Future Technology, South China University of Technology, China \\
\texttt{202364870251@mail.scut.edu.cn}
}

\begin{document}

\maketitle

\begin{abstract}
This position paper argues that LLMs should not yet be credited with decision explanation.
This matters because recent work increasingly treats accurate behavioral prediction, plausible rationales, and outcome-conditioned reasoning traces as evidence that LLMs explain why people decide as they do, risking a premature redefinition of what counts as explanatory progress in human decision modeling.
We first distinguish three claims with different evidential burdens: decision prediction, rationale generation, and decision explanation.
We then argue that the evidence most commonly offered for LLM-based decision accounts directly supports the first two claims, and sometimes explanatory hypothesis generation, but does not distinguish decision explanation from prediction-supportive rationalization.
Next, we propose a bridge standard for decision-explanation credit: stronger claims should specify explanatory targets, discriminate against weaker rationalizer alternatives, use target-appropriate process- or intervention-sensitive validation, and bound their scope.
We then situate this standard against competing views and related literatures, clarifying why it preserves the value of LLMs as predictors, narrators, and hypothesis generators while resisting premature explanatory credit.
We conclude with a principle of credit calibration: LLMs should be credited for the strongest claim their evidence warrants, and no stronger; if adopted, this principle can help turn LLMs from persuasive narrators of decisions into more reliable instruments for discovering, testing, and communicating explanations of human behavior.
\end{abstract}

\section{Introduction}

Large language models (LLMs) are increasingly used as accounts of human decisions. Given a decision context, they may be asked to predict what a person would choose, to attach a natural-language rationale to an observed or predicted choice, or to produce a reasoning trace that appears to support that choice. In related work, LLMs are also used as synthetic participants, behavioral surrogates, cognitive models, and theory-generation tools \citep{aher2023simulate,argyle2023outofone,horton2023homo,binz2024cognitive,binz2025centaur,zhu2025rlhuman}. These uses can be valuable. The mistake this paper targets is narrower: treating decision prediction, rationale generation, and decision explanation as if they licensed the same explanatory credit.

This distinction matters because decision explanation is a stronger scientific claim than fluent narration or behavioral fit. To say that an LLM explains a human decision is to say that its account is warranted as tracking some relevant structure that generated, constrained, or causally shaped the decision. If that credit is granted too easily, persuasive text can be mistaken for cognitive evidence, and progress in prediction can be misread as progress in explanation. \textbf{This position paper argues that current LLMs should not yet be credited with decision explanation.} The claim is not that LLMs can never explain decisions, but that the evidence most commonly offered today supports weaker achievements more directly than it supports that stronger credit.

The starting point is a credit ladder, developed in \S\ref{sec:claims}, that separates three claims with different evidential burdens. Decision prediction credits a model with capturing patterns in what people choose; rationale generation credits it with producing coherent or psychologically literate accounts of choices; decision explanation credits it with an account that is warranted as tracking the relevant decision-generating structure. This separation is the basis for the paper's central standard: models should be credited for the strongest claim their evidence warrants, and no stronger.

Using this ladder, \S\ref{sec:evidence} re-reads the current evidence for LLM-based decision accounts. Behavioral fit supports prediction, but does not identify which mechanism produced the behavior \citep{roberts2000goodfit,shmueli2010explain,hofman2021integrating}. Plausible rationales support rationale generation, but do not establish source tracking \citep{nisbett1977telling,jacovi2020faithful,turpin2023dontalways}. Prediction-useful traces are the strongest of the common evidence types, yet they still need not distinguish explanation from prediction-supportive rationalization \citep{wei2022cot,lanham2023measuring,paul2024making,zhu2025rlhuman}.

The positive proposal, in \S\ref{sec:requirements}, is a minimal bridge standard for decision-explanation credit. Stronger claims should specify the explanatory target, distinguish the proposed account from weaker rationalizer alternatives, use process- or intervention-sensitive validation when possible, and bound their scope by task, population, and level of analysis. This is not a complete theory of explanation; it is a discipline for matching the strength of a claim to the strength of the evidence.

Finally, \S\ref{sec:views} and \S\ref{sec:related} place the position against competing views and nearby literatures, and \S\ref{sec:conclusion} returns to its broader implication. Calibrating explanatory credit would not make LLMs less useful for human decision modeling. It would make them more useful: as predictors when they predict, as narrators when they generate rationales, as hypothesis generators when they suggest mechanisms, and eventually as explainers only when the evidence justifies that stronger role.

\section{Three Claims, Three Evidential Burdens}\label{sec:claims}

The argument begins with a simple separation: the same LLM output can be evaluated as a prediction, as a rationale, or as an explanation, but these evaluations assign different scientific credit. Table~\ref{tab:claims} summarizes the three claims and the evidence that directly bears on each.

\begin{table}[t]
\centering
\footnotesize
\caption{Three claims often compressed into ``LLM decision explanation.'' The point of the table is that each claim assigns a different kind of credit and therefore carries a different evidential burden.}
\label{tab:claims}
\begin{tabular}{@{}p{0.155\linewidth}p{0.245\linewidth}p{0.285\linewidth}p{0.195\linewidth}@{}}
\toprule
\textbf{Claim} & \textbf{Credit granted} & \textbf{Direct supporting evidence} & \textbf{What remains unlicensed} \\
\midrule
\textbf{Decision prediction} & The model captures patterns in what people choose. & Held-out or out-of-domain behavioral fit, likelihood, calibration, or choice accuracy. & That the model identifies why those choices were made. \\
\midrule
\textbf{Rationale generation} & The model produces coherent, plausible, or psychologically literate accounts of choices. & Human or expert judgments of plausibility, readability, informativeness, or theory consistency. & That the account tracks the source of the decision. \\
\midrule
\textbf{Decision explanation} & The model's account is warranted as tracking the relevant structure that generated, constrained, or causally shaped the decision. & Evidence connecting the account to a specified explanatory target and discriminating it from weaker rationalizing alternatives. & Broad credit beyond the target, task, population, and level of analysis actually tested. \\
\bottomrule
\end{tabular}
\end{table}

As Table~\ref{tab:claims} indicates, these are not three labels for the same success. Decision prediction answers whether the model gets the choice right. Rationale generation answers whether the model can produce an intelligible account of that choice. Decision explanation answers whether the account is warranted as tracking why the choice occurred at the relevant level of analysis. The evidential burden increases because the claim changes: behavioral accuracy is evidence about choices, plausible language is evidence about accounts, and explanation requires a warranted connection between an account and the decision-generating structure.

The burdens cannot be mixed because each weaker claim can be satisfied by systems that fail the stronger one. A predictor can match choices by exploiting correlations, population regularities, or task-level cues without recovering the preferences, heuristics, attention patterns, or constraints that generated them. A rationale generator can attach psychologically plausible language to a known or predicted outcome without that language tracking the decision's source. Combining the two yields a successful predictor followed by a persuasive narrator. That system may be useful, but its success does not distinguish explanation from rationalization.

This distinction fixes how positive results should be credited. Behavioral fit warrants decision-prediction credit; fluent and plausible accounts warrant rationale-generation credit; and LLM-generated candidate mechanisms may warrant credit for explanatory hypothesis generation. Decision explanation requires an additional step: evidence that links the account to an explicit explanatory target and differentiates it from weaker rationalizing alternatives. The rest of the paper evaluates current evidence against this stronger burden.

\section{What Current Evidence Actually Warrants}\label{sec:evidence}

The claim ladder in Section~\ref{sec:claims} lets us read current results without dismissing them. The issue is not that existing evidence is irrelevant to explanation. The issue is that different evidence patterns warrant different credits. In current work, three patterns recur: behavioral fit, plausible rationales, and outcome-conditioned or prediction-useful traces. Each is a real achievement. Each also leaves open a weaker interpretation than decision explanation.

\subsection{Behavioral fit warrants decision prediction}

The clearest evidence for LLM-based decision accounts is behavioral fit. Large-scale decision-modeling work has shown that machine learning can improve the prediction of human choices and can even help discover better formal models of decision-making \citep{peterson2021discover,reichman2024machine,plonsky2025predicting}. LLM-based cognitive modeling extends this ambition: models fine-tuned or evaluated on behavioral tasks can sometimes predict human choices across domains, subjects, and task descriptions with impressive accuracy \citep{binz2024cognitive,binz2025centaur}. These results deserve decision-prediction credit. They show that models can capture regularities in the mapping from decision contexts to observed choices.

But behavioral fit does not by itself say which decision-generating structure produced the choices. A risky-choice pattern may be fit by expected-value sensitivity, risk aversion, loss aversion, salience, attention, heuristics, noise, or combinations of these. More generally, a good fit can be weak evidence for a theory when many theories or flexible models can fit the same observations \citep{roberts2000goodfit}; prediction and explanation also differ in their modeling aims and validation logic \citep{shmueli2010explain,yarkoni2017choosing,hofman2021integrating}. This point is especially important for LLMs, whose behavioral similarity can be prompt-, task-, and framing-dependent. Recent work suggests that LLMs may model people as more rational than they are, may differ from cognitive process models in sequential decision tasks, and may shift strategies under external guidance \citep{liu2025rational,nguyen2024predicting,feng2025noise}. Thus, behavioral fit is evidence that a model predicts decisions. It is not yet evidence that the model explains the decision process that generated them.

\subsection{Plausible rationales warrant rationale generation}

A second common evidence pattern is the production of natural-language rationales. This achievement should not be dismissed. Explanation has a social and communicative role, and interpretable-machine-learning work has long emphasized that explanations must be evaluated relative to the task they are meant to serve \citep{miller2019explanation,doshivelez2017rigorous}. If the task is to communicate a possible reason, summarize a pattern, assist annotation, or generate hypotheses, a fluent and psychologically literate rationale may be useful. LLMs are particularly strong at this kind of reason-like text generation; for example, in computational social science tasks, free-form LLM outputs can sometimes produce explanations judged stronger than crowdworker references \citep{ziems2024css}.

The problem is that plausibility is not source tracking. Interpretability work distinguishes explanations that are plausible to readers from explanations that are faithful to the process or model being explained \citep{jacovi2020faithful}. Cognitive science gives the same warning from the human side: people may lack direct introspective access to the processes that shaped their judgments \citep{nisbett1977telling}; verbal reports can be useful data only under assumptions about how the reports were generated \citep{ericsson1980verbal}; and people can generate convincing reasons even for choices whose outcomes were covertly manipulated \citep{johansson2005failure}. The LLM case inherits this concern rather than escaping it. A model that attaches coherent psychological language to an observed or predicted choice may be generating a rationale, not identifying the source of the decision. This is why post-hoc explanation has long raised over-trust concerns in interpretable ML \citep{rudin2019stop}, and why unfaithful chain-of-thought results matter here: reason-like text can be compelling while failing to reveal the factors that actually drove a prediction \citep{turpin2023dontalways}. Plausible rationales therefore warrant rationale-generation credit, but not decision-explanation credit.

\subsection{Prediction-useful traces are the strongest weak evidence}

The strongest evidence short of decision explanation comes from intermediate traces that are tied to prediction. Chain-of-thought prompting showed that natural-language intermediate steps can improve model performance on reasoning tasks \citep{wei2022cot}. More directly for our setting, recent work has used outcome-based reinforcement learning to train LLMs to generate reasoning traces for human risky choices while also predicting those choices \citep{zhu2025rlhuman}. This is stronger than mere plausibility: the trace is not just an attractive afterthought, but part of a system optimized or evaluated in relation to behavior.

Still, a prediction-useful trace is not automatically a decision explanation. First, faithfulness to a model's own answer is itself an empirical question. Work on faithful chain-of-thought, interventions on reasoning traces, and causal mediation shows that intermediate text may or may not be what the model actually uses to produce its answer \citep{lyu2023faithful,lanham2023measuring,paul2024making,yu2026outcome}. Second, even when a trace is faithful to the model's own computation, that is not the same as being faithful to the human decision-generating structure. A trace can be optimized to support a correct prediction, to satisfy a reward signal, or to verbalize familiar psychological concepts, while still functioning as an outcome-conditioned rationalization. Recent evidence that reasoning models often do not fully verbalize the factors they use, and may sometimes settle on answers before producing much of the subsequent reasoning, reinforces the need to test rather than assume trace faithfulness \citep{chen2025reasoning,datta2026decideearly}.

The current evidence is therefore substantial but over-credited. Behavioral fit supports decision prediction; plausible rationales support rationale generation; prediction-useful traces support a stronger but still limited claim that traces may help organize prediction or generate explanatory hypotheses. This intermediate credit is important: an LLM may propose a useful mechanism, such as loss sensitivity, attention, or heuristic use, without yet having explained the decision. The problem is that many evaluations make these achievements move together: the same result may show that a model predicts a choice, can narrate that choice, and can use the narration to organize prediction. Such a result is valuable, but it is not yet diagnostic of the strongest claim. Explanatory hypothesis generation becomes decision explanation only when the proposed account is tested against evidence that is sensitive to the proposed target and separates it from plausible rationalizers. The next question is therefore not whether LLM accounts are useful, but when that usefulness can be upgraded into explanatory credit. The answer requires bridge evidence: evidence that connects a model's account to an explicit explanatory target and discriminates it from weaker rationalizing alternatives.

\section{What Decision Explanation Would Require}\label{sec:requirements}

The previous section leaves a positive task: to say when an LLM account should be upgraded from prediction, rationale generation, or explanatory hypothesis generation to decision explanation. That upgrade cannot be earned by adding prediction, plausibility, and useful traces together. Those achievements are relevant, but they do not by themselves bridge the gap between an account of a decision and the structure that generated or shaped that decision. The bridge standard developed here follows from that gap. It has four connected parts. First, the claim must specify the explanatory target: otherwise there is no determinate claim to evaluate (\S\ref{sec:target}). Second, the evidence must discriminate the proposed account from weaker rationalizers: otherwise the same result supports weaker and stronger claims at once (\S\ref{sec:rationalizers}). Third, the validation must be appropriate to the target rather than to explanation in the abstract (\S\ref{sec:bridge-evidence}). Finally, any upgraded credit must be scoped and reported so that local explanatory success is not generalized into a broader claim than the evidence supports (\S\ref{sec:scope-report}).

This is not a complete philosophical theory of explanation. Our use of ``tracking'' is deliberately pluralist, but not permissive. Different decision sciences may cash out explanation in mechanistic, causal, interventionist, model-based, or construct-validity terms \citep{woodward2003making,pearl2009causality,peters2016causal}. The bridge standard does not choose among those theories. It states a weaker requirement common to them: whatever explanatory level is claimed, the work must make the claim evaluable and make the evidence diagnostic. This is also why we do not propose universal quantitative thresholds for decision explanation. A gaze-alignment measure may be appropriate for an attention claim in visual choice, but irrelevant to an archival study of policy decisions; a parameter-recovery test may be natural for risky choice, but not for moral judgment or multi-agent bargaining. What should be standardized is not a single score for explanation, but the burden of reporting: metric selection, not metric universalization.

\subsection{Specify the explanatory target}\label{sec:target}

The first requirement is target specification. This is not a terminological preference; it is what turns ``the LLM explains the decision'' into a claim that can be evaluated. Without a target, there is no way to say what the model account is supposed to be right about, no way to choose evidence that would support or disconfirm it, and no way to distinguish a successful explanation from a plausible story. At minimum, the target may be one of the following:

\begin{description}
    \item[\textbf{Preference or parameter target}] such as loss sensitivity or temporal discounting.
    \item[\textbf{Rule or strategy target}] such as expected-value comparison or lexicographic choice.
    \item[\textbf{Information-use target}] such as attention to particular attributes or outcomes.
    \item[\textbf{Process target}] such as search order, response time, evidence accumulation, or memory limitation.
    \item[\textbf{Contextual-causal target}] such as framing, reference points, time pressure, or social cues.
    \item[\textbf{Population-structure target}] such as latent decision types or subgroup heterogeneity.
\end{description}

These targets are not interchangeable. An account that tracks the payoff feature a person attended to is different from an account that tracks the utility function they optimized; both are different from an account that tracks the social norm that made a response acceptable. Each target also induces a different evidence standard. If the claim is about attention, then gaze, information search, response dynamics, or counterfactual masking of information may be relevant. If the claim is about loss sensitivity, then changed reference points or altered gain--loss framing become more diagnostic. If the claim is about a decision heuristic, then behavior in cases that separate that heuristic from nearby alternatives is needed.

The target therefore fixes the form of the explanatory claim. It prevents the evaluation from collapsing back into generic behavioral fit or generic plausibility, where a model can sound explanatory without anyone being able to say what exactly has been explained. It also prevents post-hoc broadening: a rationale that mentions loss aversion, attention, and social norms should not be credited as explaining whichever of those later happens to fit the data. Target specification is the first step in upgrading from a plausible rationale to a decision-explanation claim because it makes the claim precise enough to bear evidence.

Target specification does not require pretending that decisions have a single cause. A claim may target several structures, but it must say whether they are competing alternatives, joint contributors, mediating steps, or sources of heterogeneity; otherwise, adding more psychological terms broadens the narrative without making the claim more testable. The point is not to simplify the psychology, but to make clear which relation among plausible targets the evidence is supposed to support before explanatory credit is assigned to the model account rather than to a looser narrative that names all plausible mechanisms.

\subsection{Discriminate against weaker rationalizer alternatives}\label{sec:rationalizers}

Once the target is specified, the evidence must be diagnostic among the claims in Table~\ref{tab:claims}. A design that simultaneously supports prediction, rationale generation, and decision explanation has not by itself supported the strongest of the three. It has left the evidential roles entangled. The relevant question is not whether the LLM did something useful, but whether the result would still hold for weaker systems that predict or narrate without tracking the stated target. This is the same methodological problem that arises when a single experimental manipulation moves several explanatory factors at once: the result may be compatible with multiple readings, but not diagnostic for the one being credited.

Common weaker alternatives include a predictor-only model that captures choice patterns without explanation; a predictor followed by a narrator that verbalizes an observed or predicted choice; an outcome-conditioned rationale generator that writes reasons after being given the decision; a theory- or protocol-retrieval narrator that reproduces familiar psychological vocabulary, textbook explanations, or published experimental protocols; and a judge-optimized rationale generator selected for plausibility rather than source tracking. These are not strawmen. They are natural alternatives whenever the evaluation rewards behavioral fit, fluent rationales, theory-consistent language, or downstream usefulness of text. If such alternatives could pass the same evaluation, then the evaluation has not isolated decision explanation.

This discrimination burden is familiar from both interpretability and cognitive modeling. Plausible explanations can be convincing to users without being faithful to the process being explained \citep{jacovi2020faithful,rudin2019stop}; good behavioral fit can be compatible with multiple theories \citep{roberts2000goodfit}. The same logic applies here. If a post-hoc narrator given the same predicted choice can produce rationales that are equally plausible, equally theory-literate, and equally useful to a downstream predictor, then those properties do not yet support decision-explanation credit. They support the weaker claim that the system can narrate or organize predictions.

The point is not that every study must defeat every possible alternative. The point is that a decision-explanation claim should identify the weaker rivals made plausible by its own design and show how the evidence separates them. In some settings this may involve comparing native traces with post-hoc rationales generated from the same choices. In others it may involve testing whether explanations remain aligned when outcomes are withheld, swapped, or counterfactually changed. It may involve comparing LLM accounts with formal cognitive models, process models, or simpler theory-labeling systems. What matters is the asymmetry: the proposed explanation should succeed where a merely prediction-supportive rationalizer would not.

\subsection{Use target-appropriate bridge evidence}\label{sec:bridge-evidence}

The first two requirements make the claim evaluable and diagnostic. The next question is what kind of evidence can carry that burden. The guiding idea is simple: bridge evidence should vary with the explanatory target, not merely with the final choice or the fluency of the rationale. Final choices and final rationales are precisely what weaker rationalizers can imitate. Bridge evidence must therefore make the account answerable to something outside that surface pairing: a process measure, an intervention, an independent construct, a diagnostic contrast, or a source of variation that would matter if the stated target were correct.

For target-level claims, process measures and interventions are especially valuable when available. Reaction times, eye movements, information search, mouse trajectories, confidence, sequential sampling patterns, within-subject variation, and carefully interpreted verbal reports can constrain process claims when matched to the target. Cognitive science has long treated verbal reports as data whose evidential force depends on assumptions about how they are generated, not as automatic access to mental causes \citep{nisbett1977telling,ericsson1980verbal,johansson2005failure}. Similarly, counterfactual interventions can test whether the account predicts how decisions would change when the alleged structure is changed: reference-point manipulations for loss sensitivity, cue-conflict cases for lexicographic heuristics, or salience changes for attention-based accounts. The common structure is not the particular measurement, but the fact that the evidence is chosen because of the target.

The same principle applies when laboratory process measures are unavailable. In archival or observational settings, the appropriate bridge evidence may instead come from natural experiments, longitudinal variation, pre-specified counterfactual contrasts, negative controls, independently measured constructs, or convergence across datasets. In purely observational settings, the warranted claim may often remain explanatory hypothesis generation unless the design supplies some independent source of variation that discriminates the target from alternatives. This is not a demotion of such work; it is the credit ladder doing its job. When the available evidence supports prediction, rationale generation, or hypothesis generation, those achievements should be credited as such. What remains unwarranted is the stronger credit of decision explanation.

This requirement also clarifies the role of chain-of-thought and other intermediate traces. A trace that helps a model predict is relevant evidence, but the bridge question is whether the trace remains aligned with independent target-sensitive constraints. Faithfulness to the model's own computation is already nontrivial \citep{turpin2023dontalways,lanham2023measuring,paul2024making}; faithfulness to the human decision-generating structure is a further claim. Model-side faithfulness can rule out empty narration, but not human-side source tracking. Process- or intervention-sensitive validation is one way to bring the model-side account and the human-side target under the same diagnostic constraints.

Across targets, useful bridge diagnostics may include rationale- or outcome-swap stress tests, counterfactual consistency tests, recovery on known-mechanism synthetic agents, mediation analyses tied to pre-specified constructs, or alignment with independently measured process data. These diagnostics are not proposed as universal metrics or thresholds. They are examples of how a study can make the chosen evidence answer the specific explanatory target and rule out the weaker rationalizers made plausible by its design.

\paragraph{A worked claim audit: risky-choice rationales.}
Consider a representative risky-choice study in which an LLM predicts human choices in gambles and produces rationales mentioning loss aversion, risk sensitivity, or expected-value comparison \citep{zhu2025rlhuman}. The first credit is straightforward: held-out choice accuracy warrants decision prediction. The second credit is also legitimate: coherent rationales that name psychologically relevant mechanisms warrant rationale generation, and perhaps explanatory hypothesis generation if they identify mechanisms worth testing. But these credits do not yet settle the explanatory claim, because several targets remain live. A reference-dependent loss-sensitivity account is different from generic risk aversion, expected-value comparison, or attention to salient outcomes.

A decision-explanation claim in this setting would therefore have to choose the target and make the evidence diagnostic for it. If the target is loss sensitivity, the study should ask whether the account remains supported under changed reference points, altered gain--loss framing, or diagnostic gamble pairs that separate loss sensitivity from risk aversion and expected-value comparison. It should also ask whether weaker alternatives could pass the same evaluation: an outcome-conditioned narrator could verbalize loss aversion after seeing the predicted choice; a theory- or protocol-retrieval narrator could attach familiar risky-choice vocabulary; and a judge-optimized narrator could produce the most plausible-sounding account without tracking the decision source. Stronger bridge evidence might come from reference-point interventions, recovery on known-mechanism agents, or independently measured attention to gains and losses. The conclusion should then be scoped to what survived those tests. Without such bridge evidence, the warranted claim is that the LLM predicts the decision and proposes a plausible explanation. With such evidence, the claim may be upgraded to a scoped explanation of loss-sensitivity-driven choices in that setting, not generic LLM decision explanation.

\subsection{Bound and report the upgraded credit}\label{sec:scope-report}

If the first three requirements specify the claim, distinguish it from weaker alternatives, and supply target-appropriate evidence, the final question is how far the upgraded credit travels. Scope is not an administrative add-on; it is part of the evidence. A study may justify saying that an LLM explanation tracks a particular strategy in a particular risky-choice task, for a particular population, at a particular level of analysis. That does not justify the broader claim that LLMs explain decisions in general. Because decision explanation is target-relative, what counts as adequate evidence for explaining a simple forced-choice task may not count for explaining moral judgment, clinical decision-making, multi-agent bargaining, or long-horizon planning.

Table~\ref{tab:bridge-checklist} restates the bridge standard as a reporting pattern. The table is not a new metric and does not impose one experimental design. It makes explicit what must be reported before a claim can be upgraded from prediction, rationale generation, or explanatory hypothesis generation to decision explanation: the level of claim, the target, the target-appropriate evidence, the weaker alternatives, the discrimination result, and the scope of the credit.

\begin{table}[t]
\centering
\footnotesize
\caption{A reporting pattern for decision-explanation claims. The items are meant to make the evidential upgrade inspectable, not to define a universal score for explanation.}
\label{tab:bridge-checklist}
\resizebox{\linewidth}{!}{%
\begin{tabular}{@{}p{0.20\linewidth}p{0.31\linewidth}p{0.36\linewidth}@{}}
\toprule
\textbf{Report} & \textbf{Question} & \textbf{Role in the credit upgrade} \\
\midrule
Claim level & Is the work claiming prediction, rationale generation, hypothesis generation, or decision explanation? & Prevents weaker achievements from being redescribed as stronger ones. \\
\midrule
Explanatory target & What structure is the account supposed to track? & Makes the explanation answerable to a parameter, rule, process, causal factor, or population structure. \\
\midrule
Target-appropriate evidence & What diagnostic is appropriate for that target? & Requires metric selection: e.g., intervention, process measure, recovery task, mediation analysis, natural experiment, or negative control. \\
\midrule
Weaker alternatives & Which rationalizers are plausible under this design? & Identifies predictor-only, predictor-plus-narrator, outcome-conditioned, protocol-retrieval, or judge-optimized alternatives. \\
\midrule
Discrimination result & How does the evidence separate the account from those alternatives? & Determines whether the claim is upgraded or remains prediction, rationale generation, or hypothesis generation. \\
\midrule
Scope & Where does the credit apply? & Bounds the claim by task, population, domain, and level of analysis. \\
\bottomrule
\end{tabular}%
}
\end{table}

The resulting rule is simple: the upgrade from prediction, rationale generation, or explanatory hypothesis generation to decision explanation is warranted only when the account has a specified target, the evidence distinguishes it from plausible rationalizer alternatives, the validation is sensitive to that target, and the resulting claim is scoped to the tested domain. This rule need not be a negative verdict. It allows weaker and stronger successes to be credited separately: prediction when the model predicts, rationale generation when it communicates plausible accounts, hypothesis generation when it suggests mechanisms worth testing, and decision explanation when bridge evidence supports the stronger claim. The result is not less credit for LLMs, but more precise credit.

\section{Other Views, Boundary Cases, and Related Work}\label{sec:views-related}

\subsection{Other Views and Boundary Cases}\label{sec:views}

Other views on LLM decision accounts can be grouped into four broad positions. The prediction-first view holds that behavioral prediction is itself a central scientific achievement, and that highly predictive models may be more useful than less accurate theories even when their mechanisms are opaque \citep{yarkoni2017choosing,hofman2021integrating,peterson2021discover,plonsky2025predicting,reichman2024machine}. The pragmatic view treats explanation as a communicative or user-facing practice: if a rationale helps a user understand, contest, or act on a decision, then it may count as an explanation for that purpose \citep{miller2019explanation}. The parity-with-humans view notes that human verbal reports are also imperfect, selective, and sometimes post hoc, so LLM rationalization may not be uniquely disqualifying \citep{nisbett1977telling,ericsson1980verbal,johansson2005failure}. Finally, the constrained-success view holds that LLMs may help explain decisions in limited settings, especially when they are embedded in cognitive-modeling pipelines, trained on behavioral feedback, or used to generate candidate mechanisms for later testing \citep{binz2024cognitive,binz2025centaur,zhu2025rlhuman}.

The strongest opposing view is not simply that LLMs are useful predictors, which we accept, but that accurate prediction plus psychologically plausible rationale is already enough to deserve the name of explanation. Our disagreement is exactly about that upgrade. The prediction-first view is compatible with our position when it claims prediction, but not when prediction is redescribed as explanation without bridge evidence. The pragmatic view is compatible when it concerns user-facing rationales, but decision explanation in our sense is a stronger scientific credit. The parity-with-humans view supports rather than weakens our caution: cognitive science treats verbal reports as data requiring methodological constraints, not as automatic access to causes. The constrained-success view is also compatible with the position, because our claim is not that LLMs can never explain decisions; it is that current evidence often licenses prediction, rationale generation, or hypothesis generation more directly than decision explanation. The disagreement, then, is not over whether LLMs are useful for decision research, nor over whether explanation is possible in principle. It is over when the field should upgrade that usefulness into explanatory credit.

\subsection{Related Work and Positioning}\label{sec:related}

The closest related work falls into six strands. First, work on LLMs as synthetic participants, economic agents, cognitive models, and foundation models of cognition shows that LLMs can approximate human responses across many settings, and sometimes uses their verbal outputs as part of the modeling story \citep{aher2023simulate,argyle2023outofone,horton2023homo,ziems2024css,binz2024cognitive,binz2025centaur,liu2025rational,nguyen2024predicting,feng2025noise}. Second, work on machine learning for human decision prediction and computational social science has long emphasized the value of predictive evaluation while distinguishing prediction from explanation and theory discovery \citep{shmueli2010explain,yarkoni2017choosing,hofman2021integrating,peterson2021discover,reichman2024machine,plonsky2025predicting}. Third, work on interventionist, causal, and invariance-based explanation motivates the demand that explanatory claims be answerable to target-sensitive counterfactuals or changes across environments \citep{woodward2003making,pearl2009causality,peters2016causal}. Fourth, work on interpretability and chain-of-thought faithfulness shows that explanation-like text can be useful, persuasive, or even prediction-relevant without being faithful to the process it purports to expose \citep{doshivelez2017rigorous,jacovi2020faithful,rudin2019stop,wei2022cot,turpin2023dontalways,lanham2023measuring,lyu2023faithful,paul2024making,chen2025reasoning,datta2026decideearly}. Fifth, mechanistic or representation-level interpretability, causal abstraction, causal-mediation, causal representation learning, and trajectory-diagnostic work can help ask whether a model internally represents or uses variables related to a proposed account \citep{doshivelez2017rigorous,jacovi2020faithful,vig2020gender,paul2024making,geiger2023causal,scholkopf2021causal}. Sixth, classic work on verbal reports and choice blindness provides a human analogue: natural-language accounts can be informative evidence, but their evidential force depends on how they are elicited and validated \citep{nisbett1977telling,ericsson1980verbal,johansson2005failure}.

These literatures motivate our position, but none of them states the same credit-calibration argument for LLM-based decision explanation. LLM simulation and cognitive-modeling work often demonstrates behavioral fidelity or useful hypothesis generation, but does not always separate those achievements from explanatory credit for generated accounts. Human-decision prediction work clarifies why prediction matters, but it usually does not address the new role of LLM rationales and reasoning traces as apparent explanations. Causal and interventionist work motivates target-sensitive validation, but it does not by itself solve the LLM-specific problem that a model can retrieve, narrate, or optimize plausible explanations without tracking the particular decision being explained. Interpretability and chain-of-thought work supplies tools for asking whether model explanations are faithful to model behavior, but faithfulness to a model's own computation is not the same as tracking the human decision-generating structure. Model-level source tracking may contribute to bridge evidence when the claim concerns the model's source of prediction, but it is not sufficient on its own: showing that an LLM internally encodes loss-related features, for example, is not the same as showing that its account tracks the human process that generated a particular choice. Verbal-report research warns against naive rationales, but it was not formulated for LLM systems that can combine prediction, narration, and theory or protocol retrieval in a single interface. Our contribution is to connect these strands around a single claim: current LLM decision accounts should be credited according to the evidential burden they actually meet. This matters because the same systems can be powerful predictors, persuasive narrators, and useful theory-generation tools without yet deserving the stronger scientific credit of decision explanation.

\section{Conclusion}\label{sec:conclusion}

This paper has argued that current LLMs should not yet be credited with decision explanation. The argument was not that LLMs are useless for studying human decisions, nor that they can never explain them. It was that decision prediction, rationale generation, and decision explanation are different claims with different evidential burdens. Current evidence often supports the first two, and sometimes supports explanatory hypothesis generation, more directly than it supports the stronger claim that an LLM account tracks the structure that generated or shaped a human decision. We therefore proposed a bridge standard for future work: specify the explanatory target, distinguish the account from weaker rationalizing alternatives, choose validation that is sensitive to that target, and bound the scope of the claim. The importance of this position is practical as well as conceptual. If adopted, it would let the field preserve the value of LLMs as predictors, narrators, and hypothesis generators while preventing premature explanatory credit. More importantly, it would turn LLM-based decision research toward sharper tests of human behavior, making LLMs not just persuasive storytellers about decisions, but better instruments for discovering, evaluating, and communicating explanations.

The main limitation of this paper is that it offers a standard for warranting decision-explanation claims rather than a complete theory of decision explanation. We have not tried to settle which explanatory level is best for every domain, nor to prescribe one universal metric, threshold, or experimental design for all LLM-based decision studies. This restraint is deliberate: different tasks may reasonably target different structures, and the relevant diagnostics should be chosen in light of those targets. Some constrained settings may eventually provide strong enough bridge evidence for genuine explanation. Our claim is therefore scoped: absent such bridge evidence, current LLM decision accounts should be credited for the weaker achievements they establish, but not yet for decision explanation.

\section*{References}
\small
\begingroup
\renewcommand{\section}[2]{}

\endgroup

\end{document}